\documentclass{article}

\usepackage{arxiv}

\usepackage[utf8]{inputenc} 
\usepackage[T1]{fontenc}    
\usepackage{hyperref}       
\usepackage{url}            
\usepackage{booktabs}       
\usepackage{amsfonts}       
\usepackage{nicefrac}       
\usepackage{microtype}      
\usepackage{bm} 
\usepackage{amsmath} 
\usepackage{multirow}
\usepackage{here}
\usepackage[pdftex]{graphicx}

\title{ROI Regularization for Semi-supervised and Supervised Learning}

\date{} 					

\author{
  Hiroshi Kaizuka \\
  NS Solutions Corporation\\
  \texttt{kaizuka.hiroshi.3hn} \\
  \texttt{@jp.nssol.nipponsteel.com} \\
   \And
  Yasuhiro Nagasaki \\
  Texeng Solutions Corporation\\
  \texttt{nagasaki.yasuhiro.x3} \\
  \texttt{@tex-sol.com} \\
   \And
  Ryo Sako \\
  University of Tsukuba\\
  \texttt{s1820454@s.tsukuba.ac.jp} \\
}

\begin{document}
\maketitle

\begin{abstract}
We propose ROI regularization (ROIreg) as a semi-supervised learning method for image classification. ROIreg focuses on the maximum probability of a posterior probability distribution $g(\bm{x})$ obtained when inputting an unlabeled data sample $\bm{x}$ into a convolutional neural network (CNN). ROIreg divides the pixel set of $\bm{x}$ into multiple blocks and evaluates, for each block, its contribution to the maximum probability. A masked data sample $\bm{x}_{\mathrm{ROI}}$ is generated by replacing blocks with relatively small degrees of contribution with random images. Then, ROIreg trains CNN so that $g(\bm{x}_{\mathrm{ROI} }
)$ does not change as much as possible from  $g(\bm{x})$. Therefore, ROIreg can be said to refine the classification ability of CNN more. On the other hand, Virtual Adverserial Training (VAT), which is an excellent semi-supervised learning method, generates data sample $\bm{x}_{\mathrm{VAT}}$ by perturbing $\bm{x}$ in the direction in which  $g(\bm{x})$ changes most. Then, VAT trains CNN so that $g(\bm{x}_{\mathrm{VAT}} )$ does not change from $g(\bm{x})$ as much as possible. Therefore, VAT can be said to be a method to improve CNN's weakness. Thus, ROIreg and VAT have complementary training effects. In fact, the combination of VAT and ROIreg improves the results obtained when using VAT or ROIreg alone. This combination also improves the state-of-the-art on "SVHN with and without data augmentation" and "CIFAR-10 without data augmentation". We also propose a method called ROI augmentation (ROIaug) as a method to apply ROIreg to data augmentation in supervised learning. However, the evaluation function used there is different from the standard cross-entropy. ROIaug improves the performance of supervised learning for both SVHN and CIFAR-10. Finally, we investigate the performance degradation of VAT and VAT+ROIreg when data samples not belonging to classification classes are included in unlabeled data.\\

{\bf Document changelog}\\
{\bf v1}  Initial release.\\

\end{abstract}


\section{Introduction}\label{sec1}
When solving the problem of classifying images into $K$ classes by convolutional neural networks (CNNs), images related to the tasks are collected as training data. Manually assigning a label (indicating the class to which the image belongs) for each image is a time consuming task when the number of training data is large. Therefore, a situation may occur where the number of unlabeled data is much larger than the number of labeled data. If the number of labeled data is not sufficient, supervised learning (SL) using only labeled data can not achieve high generalization performance. Therefore, it is expected to improve the generalization performance of CNNs by utilizing the unlabeled data existing in a large amount. Semi-supervised learning (SemiSL) is a method to realize such expectation.\\
\\[-2mm]
As one method of SemiSL, Virtual Advisual Training (VAT) \cite{15} inspired by Adversalial Training \cite{23,6} has been proposed. The method combining VAT and the "entropy minimization" term \cite{7} (this is called ENT in this paper) was the state-of-the-art for SemiSL at that time. The superiority of VAT is also confirmed in thorough performance comparison experiments on SemiSL \cite{17}. Let $g(\bm{x};\bm{\theta}_k )$ be the posterior probability distribution obtained when a data sample $\bm{x}$ is input to CNN (with weights $\bm{\theta}=\bm{\theta}_k$). VAT first finds the perturbation direction $\bm{d}$ (with unit length) of $\bm{x}$ where $g(\bm{x};\bm{\theta}_k )$ changes most sensitively. Next, VAT updates $\bm{\theta}$ so that $g(\bm{x}+\varepsilon \bm{d};\bm{\theta})$ and $g(\bm{x};\bm{\theta}_k )$ do not change as much as possible for a small positive number $\varepsilon$. Therefore, VAT can be interpreted as effectively enhancing CNN's performance by improving the weakening point where CNN's generalization performance is most deteriorated. In this sense, VAT can be said to be a learning method that competes with SL.\\
\\[-2mm]
We propose a new learning method {\em ROI Regularization (ROIreg)} that can complement the learning effect of VAT. The aim is to realize SemiSL with performance superior to VAT by simultaneously using this new learning method and VAT.\\
\\[-2mm]
Let $g_{\mathrm{max}} (\bm{x};\bm{\theta}_k )$ be the maximum probability among $g(\bm{x};\bm{\theta}_k )$. First, in ROIreg, a region $\Omega$ having a relatively small contribution to  $g_{\mathrm{max}} (\bm{x};\bm{\theta}_k )$ is extracted from the pixel region of  $\bm{x}$. Next, the masked image $\bm{x}(\Omega)$ is obtained by replacing the pixel value of the pixel included in the region  $\Omega$ with a random number. Then, ROIreg updates $\bm{\theta}$ so that $g(\bm{x}(\Omega);\bm{\theta})$ does not change as much as possible from $g(\bm{x};\bm{\theta}_k )$. Here, the complementary set $\Omega^\mathrm{c}$ of  $\Omega$ corresponds to the region of interest (ROI) which is an important area for classification. For pixels contained in $\Omega^\mathrm{c}$, $\bm{x}(\Omega)$ and $\bm{x}$ have the same pixel value. Therefore, it can be said that the ROIreg is a learning method that ensures that classification ability of CNN does not deteriorate even if input information is limited only to information included in  $\Omega^\mathrm{c}$. On the other hand, the reliability of $\Omega^\mathrm{c}$ increases according to the progress of SL using labeled data. Therefore, ROIreg has the effect of refining the features acquired by SL. In this sense, ROIreg can be said to be a learning method cooperating with SL. Therefore, by combining ROIreg which is cooperative with SL and VAT which is competitive with SL, there is a high possibility that each other can complement the learning effect. In fact, VAT+ROIreg+ENT outperforms VAT+ENT. It also improves the state-of-the-art on "SVHN with and without data augmentation" and "CIFAR-10 without data augmentation".\\
\\[-2mm]
VAT and ROIreg can also be applied to situations where only labeled data is used. We experiment in this situation whether they can improve the performance of normal SL. In this case, since SL is used, ENT is not used. \\
\\[-2mm]
In addition, we propose a data augmentation method called ROI augmentation (ROIaug) as one variation of ROIreg. However, the evaluation function used there is different from the standard cross-entropy. \\
\\[-2mm]
Finally, we investigate the issue of Class Distribution Mismatch raised in \cite{17}. In the conventional SemiSL problem setup, an unlabeled data set is created by discarding the label information of the data samples contained in the labeled data set.  As a result, any sample data in the unlabeled data set belongs to one of the $K$ classes, which is a problem setting advantageous for SemiSL. However, in \cite{17}, it is reported that the performance of SemiSL is greatly degraded in situations where data samples not belonging to any of the $K$ classes are included in the unlabeled data set. We evaluate the robustness of VAT and VAT+ROIreg against such a situation (Class Distribution Mismatch) using the animal 6 class classification problem raised in \cite{17}.
%
\section{Related Work}\label{sec2}
Various methods have been proposed for SemiSL. Among them, there is a group of methods called {\em consistency regularization}. In recent years, the best results for SemiSL have been realized by it. Consistency regularization defines two functions $f_{\mathrm{target}}(\bm{x};\bm{\theta})$ and $f(\bm{x};\bm{\theta})$ determined from CNN (weights $\bm{\theta}$, posterior probability distribution $g(\bm{x};\bm{\theta})$) and an unlabeled data sample $\bm{x}$. These are functions that can be assumed to be natural that the difference between the two is small. Then, $\bm{\theta}$ is optimized to minimize the sum of the evaluation function indicating the difference between $f_{\mathrm{target}}(\bm{x};\bm{\theta})$ and $f(\bm{x};\bm{\theta})$ and the conventional loss function for labeled data samples.\\
\\[-2mm]
VAT \cite{15} and our proposed method ROIreg analytically generate perturbed data $\bm{\tilde{x}}$ from $\bm{x}$. And they adopt $f_{\mathrm{target}}(\bm{x};\bm{\theta})=g(\bm{x};\bm{\theta})$ and $f(\bm{x};\bm{\theta})=g(\bm{\tilde{x}};\bm{\theta})$. While VAT adds a linear perturbation to $\bm{x}$, ROIreg uses a nonlinear perturbation that replaces the image of a subregion of $\bm{x}$ with a random image.\\
\\[-2mm]
In $\Pi$-model \cite{24}, $\bm{x}$ is input to CNN twice. In each forward propagation, two different values of $g(\bm{x};\bm{\theta})$ are obtained because the realized values of the stochastic elements (noise addition, dorpout patterns) are different. The $\Pi$-model adopts these as $f_{\mathrm{target}}(\bm{x};\bm{\theta})$ and $f(\bm{x};\bm{\theta})$. $\Pi$-model can also be viewed as a simplification of the $\Gamma$-model of the Ladder Network \cite{20}. $\Pi$-model is a method to enhance the generalization performance of CNN by two ensemble averages.\\
\\[-2mm]
In $\Pi$-model, the two dorpout patterns $\bm{\epsilon}_\mathrm{{target}}$ and $\bm{\epsilon}$ are determined randomly. However, a method has been proposed that analytically determines $\bm{\epsilon}$ so that the difference between $f_{\mathrm{target}}(\bm{x};\bm{\theta})$ and $f(\bm{x};\bm{\theta})$ is maximized among $\delta$-hyperspheres centered on  $\bm{\epsilon}_\mathrm{{target}}$. This method is called Virtual Adversalial Dropout (VAdD) \cite{18}.\\
\\[-2mm]
Temporal Ensembling \cite{12} is a method where $f(\bm{x};\bm{\theta})=g(\bm{x};\bm{\theta})$ and $f_{\mathrm{target}}(\bm{x};\bm{\theta})$ is an exponential moving average of $g(\bm{x};\bm{\theta})$ in the past epoch . Therefore, it can be said that Temporal Ensembling adopts an ensemble average over epochs that appeared in the training process. However, $f_{\mathrm{target}}(\bm{x};\bm{\theta})$ for each $\bm{x}$ is updated only once per epoch, so it takes time for temporal ensembling to be effective. The method to improve this is Mean Teacher \cite{24}. Mean Teacher uses CNN, which has an an exponential moving average of $\bm{\theta}$ obtained during training as weights, as a teacher CNN. Mean Teacher adopts $g(\bm{x};\bm{\theta})$ as $f(\bm{x};\bm{\theta})$ and a posterior probability distribution predicted by the teacher CNN as $f_{\mathrm{target}}(\bm{x};\bm{\theta})$.\\
\\[-2mm]
VAdD and Mean Teacher developed from $\Pi$-model are learning methods that complement with VAT and ROIreg. In fact, the simultaneous use of VAdD and VAT produces results that exceed the results obtained when VAdD and VAT are used alone \cite{18}.\\
\\[-2mm]
The current state-of-the-art SemiSL result for CIFAR-10  is achieved by fast-SWA \cite{1}, a modified version of Stochastic Weight Averaging (SWA) \cite{9}. This method uses $\Pi$-model or Mean Teacher. In fast-SWA, the learning rate is periodically changed in a sawtooth shape, and $\bm{\theta}$ used for the test is determined by sampling and averaging $\bm{\theta}$ obtained for each SGD update.\\
\\[-2mm]
There are Random Erasing \cite{25} and Cutout \cite{3} as a method to mask the partial area of the input data sample. However, these are data augmentation methods that randomly determine the area to be masked. On the other hand, our ROIreg analytically determines a region that is not important for class determination and sets the determined region as a region to be masked.
%
\section{ROI Regularization (ROIreg)}\label{sec3}
%
\subsection{Notation}\label{sec3.1}
$\mathcal{D}_\mathrm{L}$ and $\mathcal{D}_\mathrm{UL}$ denote a labeled data set and an unlabeled data set, respectively. $\mathrm{label}(\bm{x}) \in \{1,2,\cdots,K\}$ denotes the label of the labeled data sample $\bm{x} \in \mathcal{D}_\mathrm{L}$. One minibatch consists of a set $\mathcal{X}^{\mathrm{mb}}_{\mathrm{L}}$ consisting of $m_{\mathrm{L}}$ data samples randomly sampled from $\mathcal{D}_\mathrm{L}$ and a set $\mathcal{X}^{\mathrm{mb}}_{\mathrm{UL}}$ consisting of $m_{\mathrm{UL}}$ data samples randomly sampled from $\mathcal{D}_\mathrm{USL} = \mathcal{D}_\mathrm{L} \cup \mathcal{D}_\mathrm{UL}$. Define the following Shannon entropy and Kullback-Leibler divergence of $K$-dimensional probability distributions $p$ and $q$:
\begin{equation}
	H(p)=-\sum^{K}_{i=1}p_{i}{\log}p_{i},
\end{equation}
\begin{equation}
	D_{\mathrm{KL}}(p{\parallel}q)=-\sum^{K}_{i=1}p_{i}{\log}q_{i}-H(p).
\end{equation}
For $n$-dimensional real vector $\bm{z} \in \mathbb{R}^{n}$, $z_{i}$ denotes the $i$th element of $\bm{z}$. Also,
\begin{equation}
	\|\bm{z}\|_{p}=(|\bm{z}_{1}|^{p}+|\bm{z}_{2}|^{p}+ \cdots +|\bm{z}_{n}|^{p})^{1/p}   \notag
\end{equation}
denotes the $L^{p}$ norm of $\bm{z}$.
%
\subsection{Algorithm}\label{sec3.2}
\paragraph{Outline.}
In this section, we assume a situation where the weight parameter of the CNN to be learned is $\bm{\theta} = \bm{\theta}_{k}$ after the $k$th training finishes.\\
\\[-2mm]
For $\bm{x} \in \mathcal{D}_\mathrm{USL}$, execute a forward calculation
\begin{equation}
	\bm{x} \in \mathbb{R}^{N_{\mathrm{r}} \times N_{\mathrm{c}} \times N_{\mathrm{d}}}
	\underset{\text{data augmentation}}{\longrightarrow}
	\bm{x}_{\mathrm{aug}}(\bm{x})
	\longrightarrow
	\text{CNN$(\bm{\theta} = \bm{\theta}_{k})$ + softmax} 
	\longrightarrow g(\bm{x}_{\mathrm{aug}}(\bm{x});\bm{\theta}_{k}) \in  \mathbb{R}^{K}
	\notag
\end{equation}
using data augmentation (for example, random translation or random horizontal flipping) to obtain the posterior probability distribution $g(\bm{x}_{\mathrm{aug}}(\bm{x});\bm{\theta}_{k})$. In this paper, for ease of description, $\bm{x}_{\mathrm{aug}}(\bm{x})$ is simply expressed as $\bm{x}$.\\
\\[-2mm]
At this time, the class to which $\bm{x}$ belongs is determined to be a class corresponding to the maximum probability $g_{\mathrm{max}}(\bm{x};\bm{\theta}_{k})$ of $g(\bm{x};\bm{\theta}_{k})$. Here, it can be considered that the pixel region $\Omega(\bm{x})$ not relatively contributing to $g_{\mathrm{max}}(\bm{x};\bm{\theta}_{k})$ is not an important region for this class determination. Therefore, if we generate an image $\bm{x}_{\mathrm{ROI}}(\bm{x})$  in which the image at $\Omega(\bm{x})$ is replaced with a random image, $g(\bm{x}_{\mathrm{ROI}}(\bm{x});\bm{\theta})$ and $g(\bm{x};\bm{\theta}_{k})$ should be similar. This is because pixel values of $\bm{x}_{\mathrm{ROI}}(\bm{x})$ and $\bm{x}$ match in $\Omega(\bm{x})^{\mathrm{c}}$ which is an important region (the {\em region of interest (ROI)}) for this class determination.\\
\\[-2mm]
ROIreg is a method of learning $\bm{\theta}$ so that the difference between $g(\bm{x};\bm{\theta}_{k})$ and $g(\bm{x}_{\mathrm{ROI}}(\bm{x});\bm{\theta})$ is minimized. Therefore, it can be said that ROIreg is a learning method that attempts to maintain class determination accuracy of CNN even if information of $\bm{x}$ is limited only to information included in pixel region $\Omega(\bm{x})^{\mathrm{c}}$. From this point of view, ROIreg can be thought of as a learning method that can cooperate with SL.\\
\\[-2mm]
ROIreg uses
\begin{equation}
	L_{\mathrm{ROIreg}}=
	\frac{\rho_{\mathrm{ROI}}}{m_{\mathrm{UL}}}
	\sum_{\bm{x} \in {\mathcal{X}}^{\mathrm{mb}}_{\mathrm{UL}}}
	\{{d_{\mathrm{rel}} (g(\bm{x};\bm{\theta}_{k}))
	{\times}
	D_{\mathrm{KL}}(g(\bm{x};\bm{\theta}_{k}){\parallel}g(\bm{x}_{\mathrm{ROI}}(\bm{x});\bm{\theta}))}\}
\end{equation}
as the evaluation function. Here, $d_{\mathrm{rel}} \:  (0 \leq d_{\mathrm{rel}} \leq 1)$ is some kind of function for evaluating the reliability of class determination. The positive number $\rho_{\mathrm{ROI}}$ is a weighting parameter.
%
\paragraph{3D pixel sensitivity.}
For the maximum probability
\begin{equation}
	g_{\mathrm{max}}(\bm{x};\bm{\theta}_{k})
	= \max_{1 \leq j \leq K} g_{j}(\bm{x};\bm{\theta}_{k})
	\notag
\end{equation}
in $g(\bm{x};\bm{\theta}_{k})$, calculate 3D pixel sensitivity
\begin{equation}
	\bm{r}_{3\mathrm{D}}(\bm{x})=
	\frac{\left.\nabla _{\bm{r}}g_{\mathrm{max}}(\bm{x}+\bm{r};\bm{\theta}_{k})\right|_{\bm{r}=\bm{0}}}
	{\|\left.\nabla _{\bm{r}}g_{\mathrm{max}}(\bm{x}+\bm{r};\bm{\theta}_{k})\right|_{\bm{r}=\bm{0}}\|_{1}}
	=[\mathrm{element}(i,j,k)=r_{3\mathrm{D}}(\bm{x})(i,j,k)]
\end{equation}
using back-propagation.
%
\paragraph{2D region sensitivity.}
Set the division $\{\Omega_{q} \; (q=1,2, \cdots ,Q)\}$ of 2D pixel space $\Omega_{\mathrm{pixel}}=\{(i,j);i=1,2, \cdots ,N_{r},j=1,2, \cdots ,N_{\mathrm{c}}\}$ as follows:
\begin{equation}
	\bigcup^{Q}_{q=1}\Omega_{q}
	=\Omega_{\mathrm{pixel}}   
	\; \; \; (\Omega_{i} \cap \Omega_{j}=\phi \; (i \neq j)).
\end{equation}
We adopte a two-dimensional rectangular region as $\Omega_{q}$. The recommended $\Omega_{q}$ is a rectangular block of ${N_{\mathrm{r}}/8} \times {N_{\mathrm{c}}/8}$ size. The rectangular block size is one of the hyperparameters. At this time, 2D area sensitivity $r_{2\mathrm{D}}(\bm{x},\Omega_{m}) \; (m=1,2, \cdots ,Q)$ is calculated as follows:
\begin{equation}
	r_{2\mathrm{D}}(\bm{x},\Omega_{m})
	=\sum_{(i,j) \in \Omega_{m}}
	\sum^{N_{\mathrm{d}}}_{k=1}|r_{3\mathrm{D}}(\bm{x})(i,j,k)|.
\end{equation}
Therefore, it can be considered that the 2D region sensitivity $r_{2\mathrm{D}}(\bm{x},\Omega_{m})$ is an index of how much the pixel group included in the region $\Omega_{m}$ contributes to class determination at $\bm{\theta} = \bm{\theta}_{k}$. Note that $\sum^{Q}_{m=1}r_{2\mathrm{D}}(\bm{x},\Omega_{m}) = 1$.
%
\paragraph{Region to be masked.}
Arrange $\{r_{2\mathrm{D}}(\bm{x},\Omega_{q_{i}})  \;  (i=1,2, \cdots ,Q)\}$ by aligning 2D region sensitivities in ascending order:
\begin{equation}
	r_{2\mathrm{D}}(\bm{x},\Omega_{q_{1}})
\le \cdots \le
	r_{2\mathrm{D}}(\bm{x},\Omega_{q_{Q}})
\end{equation}
The mask replacement ratio $\lambda$ (one of hyperparameters, $0<\lambda<1$) is set, and the region $\Omega(\bm{x})$ to be masked is determined by Algorithm 1.
\begin{table}[H]
\centering
\begin{tabular}{l} 
\hline
\textbf{Algorithm 1} \; \; Method of determining the region $\Omega(\bm{x})$ to be masked.\\ 
\hline
  \hspace{5mm} Initialize $\Omega(\bm{x})=\phi$\\
  \hspace{5mm} Initialize $i=0$\\ 
  \hspace{5mm} Initialize $\mu=0$\\ 
  \hspace{5mm} \textbf{while} $\mu < \lambda$  \textbf{do}\\
      \hspace{10mm} $i \leftarrow i+1$\\
      \hspace{10mm} $\Omega(\bm{x}) \leftarrow \Omega(\bm{x}) \cup \Omega_{q_{i} } $\\ 
      \hspace{10mm} $\mu \leftarrow  \mu + r_{2\mathrm{D}}(\bm{x},\Omega_{q_{i}})$\\
   \hspace{5mm} \textbf{end while}\\
   \hspace{5mm} \textbf{return} $\Omega(\bm{x})$\\ 
\hline               
\end{tabular}
\end{table}
%
\paragraph{Masked data.}
Let $m(i,j,k)$ and $\sigma (i,j,k)$ be the average value and standard deviation of the pixel values of all the data samples included in $\mathcal{D}_\mathrm{USL}$ for each pixel $(i,j,k)$. At this time, using the random number $n(i,j,k)$ sampled from the uniform random number in the range of $[-1,1]$ independently for each pixel, masked data $\bm{x}_{\mathrm{ROI}}(\bm{x})$ is generated as follows:
\begin{equation}
	\bm{x}_{\mathrm{ROI}}(\bm{x})=  \begin{cases}
		x(i,j,k) & ((i,j) \notin \Omega(\bm{x})) \\
		m(i,j,k) + \sigma(i,j,k) \cdot n(i,j,k)  &  ((i,j) \in \Omega(\bm{x})) 
	\end{cases}
\end{equation}
%
\paragraph{Evaluation function.}
The evaluation function minimized by ROIreg is an evaluation function
\begin{equation}
	L_{\mathrm{CE}} + L_{\mathrm{VAT}} + L_{\mathrm{ROIreg}} + L_{\mathrm{ENT}} 
\end{equation}
obtained by adding the following four evaluation functions $L_{\mathrm{CE}}$,  $L_{\mathrm{VAT}}$, $L_{\mathrm{ROIreg}}$ and $L_{\mathrm{ENT}}$. $L_{\mathrm{CE}}$ is calculated using label information and the remaining three evaluation functions are calculated without using label information.
\begin{equation}
	L_{\mathrm{CE}}=
	- \frac{1}{m_{\mathrm{L}}}
	\sum_{\bm{x}\in{\mathcal{X}}^{\mathrm{mb}}_{\mathrm{L}}}
	\log g_{\mathrm{label}(\bm{x})}(\bm{x};\bm{\theta})
\end{equation}
\begin{equation}
	L_{\mathrm{VAT}}=
	\frac{1}{m_{\mathrm{UL}}}
	\sum_{\bm{x}\in{\mathcal{X}}^{\mathrm{mb}}_{\mathrm{UL}}}
	D_{\mathrm{KL}}(g(\bm{x};\bm{\theta}_{k}){\parallel}g(\bm{x} + \bm{r}_{\mathrm{vadv}}(\bm{x});\bm{\theta}))
\end{equation}
\begin{equation}
	\bm{r}_{\mathrm{vadv}}(\bm{x}) = \varepsilon \cdot \frac{\tilde{\bm{r}}} 
	{\|\tilde{\bm{r}}\|_{2}}   \notag
\end{equation}
\begin{equation}
	\tilde{\bm{r}}
	= \left.\nabla _{\bm{r}}  
	D_{\mathrm{KL}}(g(\bm{x};\bm{\theta}_{k}) \parallel g(\bm{x} + \bm{r};\bm{\theta}_{k}))
	\right|_{\bm{r}=\xi \bm{d}(\bm{x})}
    \notag
\end{equation}
\begin{equation}
	(\varepsilon>0,\xi=10^{-6},  
	\bm{d}(\bm{x}) 
	\text{ is a random vector of unit length and is independent for each $\bm{x}$.})
    \notag
\end{equation}
\begin{equation}
	L_{\mathrm{ROIreg}}=
	\frac{\rho_{\mathrm{ROI}}}{m_{\mathrm{UL}}}
	\sum_{\bm{x}\in{\mathcal{X}}^{\mathrm{mb}}_{\mathrm{UL}}}
	\{{d_{\mathrm{rel}} (g(\bm{x};\bm{\theta}_{k}))
	{\times}
	D_{\mathrm{KL}}(g(\bm{x};\bm{\theta}_{k}){\parallel}g(\bm{x}_{\mathrm{ROI}}(\bm{x});\bm{\theta}))}\}
	\; \; \; (\rho_{\mathrm{ROI}} \ge 0)
\end{equation}
\begin{equation}
	L_{\mathrm{ENT}}=
	\frac{1}{m_{\mathrm{UL}}}
	\sum_{\bm{x}\in{\mathcal{X}}^{\mathrm{mb}}_{\mathrm{UL}}}
	H(g(\bm{x};\bm{\theta}))
\end{equation}
In this paper, learning using $L_{\mathrm{CE}} + L_{\mathrm{VAT}}$ or $L_{\mathrm{CE}} + L_{\mathrm{VAT}} + L_{\mathrm{ROIreg}} + L_{\mathrm{ENT}}$ as the evaluation function is expressed as VAT or VAT+ROIreg+ENT, respectively. Notation other than these has the same meaning.\\
\\[-2mm]
The rationale of ROIreg is based on the premise that the reliability of the posterior probability distribution $g(\bm{x};\bm{\theta}_{k})$ is high, that is, the class corresponding to the maximum probability matches $\mathrm{label}(\bm{x})$. Entropy $H(g(\bm{x};\bm{\theta}_{k}))$ is a function that evaluates the degree of randomness of $g(\bm{x};\bm{\theta}_{k})$. Therefore, $H(g(\bm{x};\bm{\theta}_{k})) / \log K$ normalized to $[0,1]$ by dividing by $\log K$ can be one indicator to evaluate the uncertainty of  $g(\bm{x};\bm{\theta}_{k})$. For this reason, we evaluate the reliability $d_{\mathrm{rel}} (g(\bm{x};\bm{\theta}_{k}))$ of the class determination at $\bm{\theta} = \bm{\theta}_{k}$ by the following function:
\begin{equation}
	d_{\mathrm{rel}} (g(\bm{x};\bm{\theta}_{k}))
	= 1 - \frac{H(g(\bm{x};\bm{\theta}_{k}))}{\log K}.
\end{equation}
%
\section{Experiments}\label{sec4}
We employ training by Adam \cite{10} in all the experiments presented in this paper. Also, Adam($l_{r}$,$n_{\mathrm{update}}$,$n_{\mathrm{decay}}$) shows the following training.
\begin{quote}
In the first $n_{\mathrm{update}}-n_{\mathrm{decay}}$ updates, settings of learning rate $=l_{r}$, $\beta_{1}=0.9$ and $\beta_{2}=0.999$ are used. In the last $n_{\mathrm{decay}}$ updates, settings are used such that $\beta_{1}=0.5$ and $\beta_{2}=0.999$ and the learning rate is linearly decayed from $l_{r}$ to 0.
\end{quote}
\subsection{Comparison to Other Methods}\label{sec4.1}
\begin{figure}[H]
  \centering
  \includegraphics[keepaspectratio, scale=0.95]{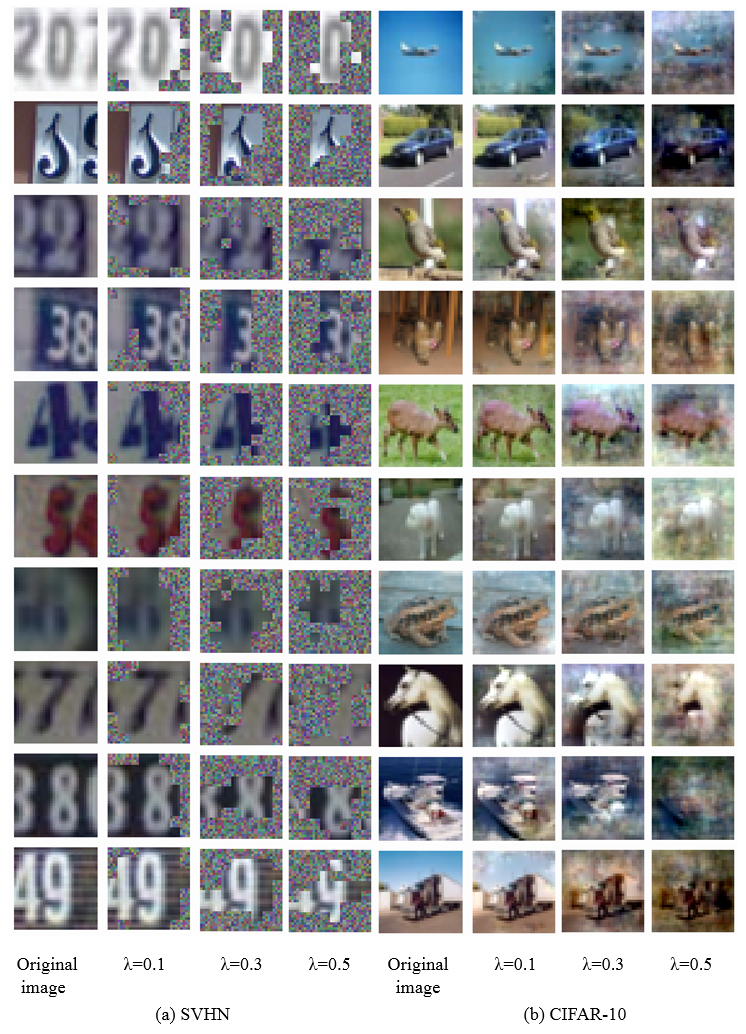}
  \caption{Examples of masked data samples $\bm{x}_\mathrm{ROI} (\bm{x})$ in CNN for which training has been completed. The CNN used in (a) achieves an error3 of 3.43\%. The CNN used in (b) achieves an error4 of 8.95\%.}
  \label{fig1}
\end{figure}

\paragraph{Datasets.}
We perform experiments using the Street View House Numbers (SVHN) dataset \cite{16} and the CIFAR-10 dataset \cite{11}. Both data sets are composed of RGB images with $32 \times 32$ pixels. SVHN is a close-up image of the house number and each image has a label corresponding to the number from 0 to 9 located at the center. CIFAR-10 consists of natural images classified into 10 classes such as airplanes, dogs and horses. In SVHN, the training set and the test set contain 73,257 images and 26,032 images, respectively. In CIFAR-10, the training set and the test set contain 50,000 images and 10,000 images, respectively.\\
\\[-2mm]
For SVHN, a 1,000 sample dataset is separated from the training set for validation. From the remainder, a 1,000 sample dataset is taken as $\mathcal{D}_\mathrm{L}$ and the remaining data set is taken as $\mathcal{D}_\mathrm{UL}$. As preprocessing, image data samples are linearly transformed to floating point values in the range $[-1,1]$. As data augmentation, we use only random translation by up to 2 pixels.\\
\\[-2mm]
For CIFAR-10, a 1,000 sample dataset is separated from the training set for validation. From the remainder, a 4,000 sample dataset is taken as $\mathcal{D}_\mathrm{L}$ and the remaining data set is taken as $\mathcal{D}_\mathrm{UL}$. As preprocessing, ZCA normalization \cite{11} is applied to the image data samples using the statistics calculated for the training set. As data augmentation, we use random horizontal flipping and random translation by up to 2 pixels.
\paragraph{CNN and Training.}
Our main purpose is to demonstrate how much VAT+ROIreg+ENT outperforms VAT+ENT. Therefore, we adopt the CNN (Conv-Large model) and learning schedule used in \cite{15} as they are. That is, as CNN, we use CNN with almost the same architecture as 13-layer CNN (with 3.1 M parameters) proposed in \cite{12}. Also, our CNN does not include weight normalization \cite{22}. As learning schedule, Adam (0.001, 48000, 16000) is used for SVHN and Adam (0.001, 200000, 16000) is used for CIFAR-10. We measure the test error rate for the CNN obtained when the last update is completed. Therefore, we do not use any early sropping method. The only difference from \cite{15} is the minibatch configuration. We use $(m_{\mathrm{L}},m_{\mathrm{UL}} )=(32,128)$ if we do not use data augmentation and $(m_{\mathrm{L}},m_{\mathrm{UL}} )=(64,96)$ if we use data augmentation. In \cite{15}, $(m_{\mathrm{L}},m_{\mathrm{UL}} )=(32,128)$ is always adopted. The details of the CNN we use are described in appendix \ref{appA}.
\paragraph{Classification Results on SVHN and CIFAR-10.}
Table \ref{tab1} shows the hyperparameter values used in our experiments. The value of the hyperparameter $\varepsilon$ for VAT is the value recommended in \cite{14}. For hyperparameters on ROIreg, we only tune when using data augmentation (see section \ref{sec4.2}). The value decided there is also applied to the case where data augmentation is not used.\\
\begin{table}[H]
\centering
\begin{tabular}{l|ccc|c}
\hline
\multicolumn{1}{c|}{\multirow{2}{*}{Dataset}} & \multicolumn{3}{c|}{ROIreg} & VAT \\ \cline{2-5} 
\multicolumn{1}{c|}{}                         & $\rho_\mathrm{ROI}$ & $\lambda$ & $\Omega_q$ & $\varepsilon$ \\ \hline
                                       SVHN+ & 0.9 & 0.5 & 4$\times$4 & 3.5 \\
                                       SVHN     & 0.9 & 0.5 & 4$\times$4 & 2.5 \\
                                       CIFAR-10+     & 1.5  & 0.5 & 4$\times$4 & 8.0  \\
                                       CIFAR-10     & 1.5  & 0.5 & 4$\times$4 & 10.0  \\ \hline
                                       
\end{tabular}
\vspace{5mm}
  \caption{Hyperparameter settings used in our experiments. "+" indicates data augmentation.}
  \label{tab1}
\end{table}
Table \ref{tab2} shows the experimental results. VAT+ROIreg+ ENT outperforms VAT+ENT and ROIreg+ENT in any case. These results demonstrate our initial expectations. In other words, ROIreg, which is learning that cooperates with SL, and VAT, which is learning that competes with SL, have the ability to mutually complement the learning effect and achieve high performance.\\
\\[-2mm]
Also, VAT+ROIreg+ENT achieves the state-of-the-art performance except for CIFAR-10 where data augmentation is used (denoted as CIFAR-10+). However, for CIFAR-10+, the test error rate of 9.33\% achieved by VAT+ROIreg+ENT does not reach the result of 9.22\% achieved by VAdD(QE)+VAT+ENT \cite{18} and the result of 9.05\% achieved by MT+fast-SWA \cite{1}. As the cause of this, it is possible that Weight Normalizaton \cite{22} used in the experiment of VAdD  and fast-SWA is not used in our experiment. Section \ref{sec5.3} describes the experiment when Weight Normalizaton is applied. Here, in experiments with SVHN+ and SVHN in VAdD, Weight Normalizaton is not applied.\\
\\[-2mm]
Figure \ref{fig1} shows examples of masked data samples $\bm{x}_{\mathrm{ROI}}(\bm{x})$. It can be seen that $\bm{x}_{\mathrm{ROI}}(\bm{x})$ deviates from the input data sample $\bm{x}$ as the mask replacement ratio $\lambda$ increases. In the case of CIFAR-10, $\bm{x}_{\mathrm{ROI}}(\bm{x})$ is generated for the data sample that has been preprocessed by ZCA on $\bm{x}$. Therefore, as shown in Figure 1 (b), when $\bm{x}_{\mathrm{ROI}}(\bm{x})$ is converted back to the original image representation, $\bm{x}_{\mathrm{ROI}}(\bm{x})$ and $\bm{x}$ are images that have different values in all pixels.
\newpage
\begin{table}[h]
  \centering
\begin{tabular}{llcc}
\hline
 & \multicolumn{1}{c}{\multirow{3}{*}{Method}}       & \multicolumn{2}{c}{Test error rates (\%)}                                               \\ \cline{3-4} 
 & \multicolumn{1}{c}{}                              & SVHN                                      & CIFAR-10                                    \\
 & \multicolumn{1}{c}{}                              & 1k labels                                 & 4k labels                                   \\ \hline
\multicolumn{4}{l}{On Conv-Large used in \cite{12}, With data augmentation}                                                   \\
 & Supervised-only \cite{24}        & 12.32 $\pm$ 0.95                          & 20.66 $\pm$ 0.57                            \\
 & Mean Teacher (MT) \cite{24}      & 3.95 $\pm$ 0.19                           & 12.31 $\pm$ 0.28                            \\
 & VAT+ENT \cite{15}                & 3.86                                      & 10.55                                       \\
 & VAdD(QE) \cite{18}               & 4.26 $\pm$ 0.14                           & 11.32 $\pm$ 0.11                            \\
 & VAdD(QE)+VAT+ENT  \cite{18}      & 3.55 $\pm$ 0.07                           & 9.22 $\pm$ 0.10                             \\
 & MT+fast-SWA   \cite{1}           &                                           & \textbf{9.05 $\pm$ 0.21}   \\
 & Ours: ROIreg+ENT                                  & 4.63 $\pm$ 0.21                           & 12.94 $\pm$ 0.29                            \\
 & Ours: VAT+ROIreg+ENT                              & \textbf{3.44 $\pm$ 0.22} & 9.33 $\pm$ 0.21                             \\
 & Ours: VAT+ROIreg+ENT                              &                                           & \: 9.13 $\pm$ 0.20*                            \\ \hline
\multicolumn{4}{l}{On Conv-Large used in \cite{12}, Without data augmentation}                                                \\
 & Supervised-only  \cite{24}       & 14.15 $\pm$ 0.87                          & 24.47 $\pm$ 0.50                            \\
 & Mean Teacher  \cite{24}          & 5.21 $\pm$ 0.21                           & 17.74 $\pm$ 0.30                            \\
 & VAT+ENT  \cite{15}               & 4.28                                      & 13.15                                       \\
 & Ours: ROIreg+ENT                                  & 4.57 $\pm$ 0.07                           & 16.70 $\pm$ 0.44                            \\
 & Ours: VAT+ROIreg+ENT                              & \textbf{3.69 $\pm$ 0.19} & 12.44 $\pm$ 0.22                            \\
 & Ours: VAT+ROIreg+ENT                              &                                           & \: \textbf{12.20 $\pm$ 0.16*} \\ \hline
\multicolumn{4}{l}{On Conv-Small used in \cite{21}, Without data augmentation}                                                \\
 & GAN (feature matching) \cite{21} & 8.11 $\pm$ 1.30                           & 18.63 $\pm$ 2.32                            \\
 & bad GAN \cite{2}                 & 4.25 $\pm$ 0.03                           & 14.41 $\pm$ 0.30                            \\ \hline
\end{tabular}\vspace{5mm}
  \caption{Test error rates (\%) on SVHN and CIFAR-10. Error bars correspond to the standard deviation over 5 runs. $\mathcal{D}_\mathrm{L}$ is chosen randomly for each experiment. Our results show error3 for SVHN+, SVHN and CIFAR-10, and error4 for CIFAR-10+. * shows the values of error2 obtained in the experiment described in section \ref{sec5.3}. Error2, error3 and error4 are the error types defined in section \ref{sec4.3}. "+" indicates data augmentation.}
  \label{tab2}
\end{table}
\newpage
\newpage
\subsection{Hyperparameters}\label{sec4.2}
ROIreg has three hyperparameters listed in Table \ref{tab1}, ie, the size of $\Omega_{q}$, the mask replacement ratio $\lambda$ and the weight $\rho_{\mathrm{ROI}}$. Patterns such as edges are important for image characterization. Therefore,  $\Omega_{q}$ of $1 \times 1$ size (division in pixel units) is inappropriate. In this paper, we adopt ${N_{\mathrm{r}}/8} \times {N_{\mathrm{c}}/8}$ size. Thus, the hyperparameters that need to be tuned are $\lambda$ and $\rho_{\mathrm{ROI}}$. Table 3 shows the hyperparameter tuning results for using data augmentation. The optimal $\lambda$ for both SVHN+ and CIFAR-10+ is 0.5. However, the optimal $\rho_{\mathrm{ROI}}$ differs between the two.\\
\\[-2mm]
\begin{table}[H]
  \centering
\begin{tabular}{c|c|ccc}
\hline
\multicolumn{2}{c|}{\multirow{2}{*}{SVHN+}} & \multicolumn{3}{c}{$\lambda$}           \\ \cline{3-5} 
\multicolumn{2}{c|}{}                       & 0.4         & 0.5         & 0.6         \\ \hline
\multirow{4}{*}{$\rho_\mathrm{ROI}$}     & 0.8     &             & 3.65 $\pm$ 0.25 &             \\
                                  & 0.9     & 3.86 $\pm$ 0.14 & 3.44 $\pm$ 0.22 & 4.04 $\pm$ 0.18 \\
                                  & 1.0     &             & 3.69 $\pm$ 0.12 &             \\
                                  & 1.5     &             & 3.88 $\pm$ 0.36 &             \\ \hline
\end{tabular}
\end{table}
%
\begin{table}[h]
\centering
\begin{tabular}{c|c|ccc}
\hline
\multicolumn{2}{c|}{\multirow{2}{*}{CIFAR-10+}} & \multicolumn{3}{c}{$\lambda$}           \\ \cline{3-5} 
\multicolumn{2}{c|}{}                           & 0.4         & 0.5         & 0.6         \\ \hline
\multirow{4}{*}{$\rho_\mathrm{ROI}$}       & 1.0       &             & 9.84 $\pm$ 0.09 &             \\
                                    & 1.5       & 9.67 $\pm$ 0.29 & 9.33 $\pm$ 0.21 & 9.65 $\pm$ 0.27 \\
                                    & 1.6       &             & 9.46 $\pm$ 0.19 &             \\
                                    & 2.0       &             & 9.71 $\pm$ 0.19 &             \\ \hline
\end{tabular}
\vspace{5mm}
  \caption{Test error rates when hyperparameters are changed. The experimental results show test error rates, where error3 is shown for SVHN+ and error4 for CIFAR-10+. Error3 and error4 are the error types defined in section \ref{sec4.3}. Error bars correspond to the standard deviation over 5 runs.}
  \label{tab3}
\end{table}
When applying VAT+ROIreg+ENT to a new data set, first tune $\varepsilon$ for VAT+ENT to find the optimal $\varepsilon_{\mathrm{opt}}$. Next, tune $\rho_{\mathrm{ROI}}$ for VAT+ROIreg+ENT where $\varepsilon = \varepsilon_{\mathrm{opt}}$, the size of $\Omega_{q}={N_{\mathrm{r}}/8} \times {N_{\mathrm{c}}/8}$  and $\lambda=0.5$. Such two-step tuning may be an efficient means.
%
\subsection{Batch Normalization Statistics}\label{sec4.3}
\paragraph{Motivation.}
In situations where CNN is actually applied, it is necessary to determine the mean and standard deviation required by batch normalization \cite{8}. Usually, batch normalization is calculated using those statistics (called BN statistics) acquired during the training period. In the case of supervised learning, this method has no problem. However, VAT and ROIreg use data samples for training that differ considerably from the input data samples. Therefore, how to determine BN statistics is a problem that greatly affects the performance of CNN trained with VAT and ROIreg. For example, updating the BN statistics with the perturbed data sample $\bm{x}_{\mathrm{VAT}}(\bm{x})$ generated by VAT will make the test error rate worse. The reason is considered to be that $\bm{x}_{\mathrm{VAT}}(\bm{x})$ and  $\bm{x}$ are quite different from the viewpoint of classification. This is because $\bm{x}_{\mathrm{VAT}}(\bm{x})$ is a data sample in which the area of $\bm{x}$ important to classification is corrupted.\\
\\[-2mm]
In the following description, CNN${}_0$ indicates the CNN obtained at the end of training. In addition, CNN${}_f$ indicates a CNN in which BN statistics have been updated by forward propagation of intentionally designed minibatches to CNN${}_0$. We use CNN${}_f$ rather than CNN${}_0$ at test time. Since we update BN statistics by  $\hat{x}_t=0.9 \times \hat{x}_{t-1}+0.1 \times x_t$, if we propagate 60 minibatches forward to CNN${}_0$, the BN statistics acquired during training will be completely renewed. 
%
\paragraph{Error Type Definition.}
Define the following three test error rates:
\begin{itemize}
	\item Error2: One minibatch is constructed by 128 data samples $\{\bm{x}_{i} \}$ randomly sampled from $\mathcal{D}_\mathrm{L}$. Error2 is defined as the test error rate of CNN${}_f$ obtained by using 60 minibatches configured in this way.
	\item Error3: One minibatch${}_a$ is constructed by data sample $\{\bm{x}_{i} \}$. At the same time, one minibatch${}_b$ is constructed by 128 masked data samples $\{\bm{x}_{\mathrm{ROI}} (\bm{x}_i )\}$ generated from $\{\bm{x}_{i} \}$. Error3 is defined as the test error rate of CNN${}_f$ obtained using 30 sets of minibatch${}_a$ and minibatch${}_b$ configured in this way. Here, forward propagation is performed 60 times.
	\item Error4: Apply data augmentation to data sample $\{\bm{x}_{i} \}$ to construct one minibatch. Error4 is defined as the test error rate of CNN${}_f$ obtained by using 60 minibatches configured in this way.
\end{itemize}
%
\paragraph{Results.}
Table \ref{tab4} shows the experimental results. For SVHN+ and SVHN, error2 and error4 that do not include updating of BN statistics by  $\{\bm{x}_{\mathrm{ROI}} (\bm{x} )\}$ are obviously larger than error3 that includes them. Such superiority of error3 against error2 is maintained in the case of CIFAR-10, although the difference is reduced. The reason for the superiority of error3 is considered to be that $\bm{x}_{\mathrm{ROI}} (\bm{x} )$ holds the pure information necessary for classification. However, in the case of CIFAR-10+, conversely, error3 shows the maximum test error rate.\\
\begin{table}[H]
  \centering
\begin{tabular}{ccc}
\hline
Test error & SVHN+                & CIFAR-10+            \\
rates      & 1k labels            & 4k labels            \\ \hline
error2     & 3.71 $\pm$ 0.28          & 9.43 $\pm$ 0.21          \\
error3     & \textbf{3.44 $\pm$ 0.22} & 9.52 $\pm$ 0.23          \\
error4     & 3.74 $\pm$ 0.30          & \textbf{9.33 $\pm$ 0.21} \\ \hline
\end{tabular}
\end{table}
%
\begin{table}[H]
\centering
\begin{tabular}{ccc}
\hline
Test error & SVHN                 & CIFAR-10              \\
rates      & 1k labels            & 4k labels             \\ \hline
error2     & 4.10 $\pm$ 0.22          & 12.54 $\pm$ 0.31          \\
error3     & \textbf{3.69 $\pm$ 0.19} & \textbf{12.44 $\pm$ 0.22} \\ \hline
\end{tabular}
\vspace{5mm}
  \caption{Comparison of test error rate (\%) among error types in the experimental results of VAT+ROIreg+ENT listed in Table \ref{tab2}. Error bars correspond to the standard deviation over 5 runs.}
  \label{tab4}
\end{table}
Let us consider the reason for such a reversal phenomenon for error3. In the case of SVHN, the region in $\bm{x}$ that is most important for classification is left in $\bm{x}_{\mathrm{ROI}} (\bm{x} )$. On the other hand, in the case of CIFAR-10, $\bm{x}_{\mathrm{ROI}} (\bm{x} )$ is generated by masking a partial region of the ZCA-processed data sample. Therefore, the data sample that can be generated by performing ZCA inverse transformation on $\bm{x}_{\mathrm{ROI}} (\bm{x} )$ has different values from $\bm{x}$ at all pixels. Because of these differences, the information purity of $\bm{x}_{\mathrm{ROI}} (\bm{x} )$ can be interpreted as lower in CIFAR-10 than in SVHN. In particular, in the case of CIFAR-10+, the presence of data augmentation will further reduce the information purity of $\bm{x}_{\mathrm{ROI}} (\bm{x} )$. This difference in information purity is considered to be one factor of the reversal phenomenon for error3.\\
\\[-2mm]
We propose to select the error type as follows. Basically, use error3. However, if you use preprocessing (e.g. ZCA) that changes the correlation of all pixels, and also use data augmentation, use error2 or error4 instead of error3. If you use preprocessing that changes the correlation of all pixels, and do not use data augmentation, use error2 or error3. 
%
\section{Discussion}\label{sec5}
\subsection{Supervised Learning: ROI Augmentation}\label{sec5.1}
VAT and ROIreg are methods for SemiSL. However, we can apply VAT and ROIreg to training using only labeled data by setting $\mathcal{D}_\mathrm{USL} = \mathcal{D}_\mathrm{L}$ and $\mathcal{X}^{\mathrm{mb}}_{\mathrm{UL}} = \mathcal{X}^{\mathrm{mb}}_{\mathrm{L}}$.\\
\\[-2mm]
Table \ref{tab5} shows the experimental results. For SVHN+, neither VAT nor ROIreg can improve the results obtained by supervised learning using $L_{\mathrm{CE}}$ alone. On the other hand, for CIFAR-10+, both VAT and VAT+ROIreg improve the results of supervised learning.\\
\\[-2mm]
By performing the following replacement
\begin{equation}
	d_{\mathrm{rel}} (g(\bm{x};\bm{\theta}_{k}))
	= 1 - \frac{H(g(\bm{x};\bm{\theta}_{k}))}{\log K}
	\rightarrow
	g_{\mathrm{label}(\bm{x})}(\bm{x};\bm{\theta}_{k})
\end{equation}
\begin{equation}
	D_{\mathrm{KL}}(g(\bm{x};\bm{\theta}_{k}){\parallel}g(\bm{x}_{\mathrm{ROI}}(\bm{x});\bm{\theta}))
	\rightarrow
	-\log g_{\mathrm{label}(\bm{x})}(\bm{x}_{\mathrm{ROI}}(\bm{x});\bm{\theta})
\end{equation}
in $L_{\mathrm{ROIreg}}$, the evaluation function
\begin{equation}
	L_{\mathrm{ROI}}^{\mathrm{aug}}
	= - \frac{\rho_{\mathrm{ROI}}}{m_{\mathrm{L}}}
	\sum_{\bm{x}\in{\mathcal{X}}^{\mathrm{mb}}_{\mathrm{L}}}
	\{g_{\mathrm{label}(\bm{x})}(\bm{x};\bm{\theta}_{k})
	{\times}
	\log g_{\mathrm{label}(\bm{x})}(\bm{x}_{\mathrm{ROI}}(\bm{x});\bm{\theta})\}
\end{equation}
is obtained. ROIreg can be used as a method for data augmentation if $L_{\mathrm{CE}}+L_{\mathrm{ROI}}^{\mathrm{aug}}$ is adopted as the evaluation function. We refer to this method as {\em ROI augmentation (ROIaug)}. Here, the mask replacement ratio $\lambda$ used in ROIaug is set to about $1/10$ of $\lambda$ in the case of ROIreg. Table \ref{tab5} shows the experimental results for ROIaug. ROIaug improves supervised learning results for both SVHN+ and CIFAR-10+.
\begin{table}[h]
  \centering
\begin{tabular}{lcc}
\hline
\multicolumn{1}{c}{\multirow{2}{*}{Method}} & SVHN+           & CIFAR-10+       \\
\multicolumn{1}{c}{}                        & 1k labels       & 4k labels       \\ \hline
Supervised-only                             & 11.00 $\pm$ 0.98    & 26.03 $\pm$ 0.57    \\
VAT                                         & 18.11 $\pm$ 0.95    & 20.19 $\pm$ 0.96    \\
ROIreg                                      & 11.54 $\pm$ 0.68    &\\
VAT+ROIreg                                  &                 & 19.61 $\pm$ 0.26    \\ \hline
\multirow{2}{*}{ROIaug}                     & 9.92 $\pm$ 0.93     & 24.83 $\pm$ 0.67    \\
                                            & $\lambda$= 0.03 & $\lambda$= 0.05 \\ \hline
\vspace{5mm}
\end{tabular}
  \caption{Test error rates (error 4) (\%) when using only labeled data. Error bars correspond to the standard deviation  over 5 runs. We use Adam (0.001, 4800, 2400) and Adam (0.0001, 24000, 12000) for SVHN+ and CIFAR-10+, respectively. The minibatch size is 100. $\mathcal{D}_\mathrm{L}$ is chosen randomly for each experiment.}
  \label{tab5}
\end{table}
%
\subsection{Class Distribution Mismatch \cite{17}}\label{sec5.2}
\begin{figure}[H]
  \centering
  \includegraphics[keepaspectratio, scale=0.7]{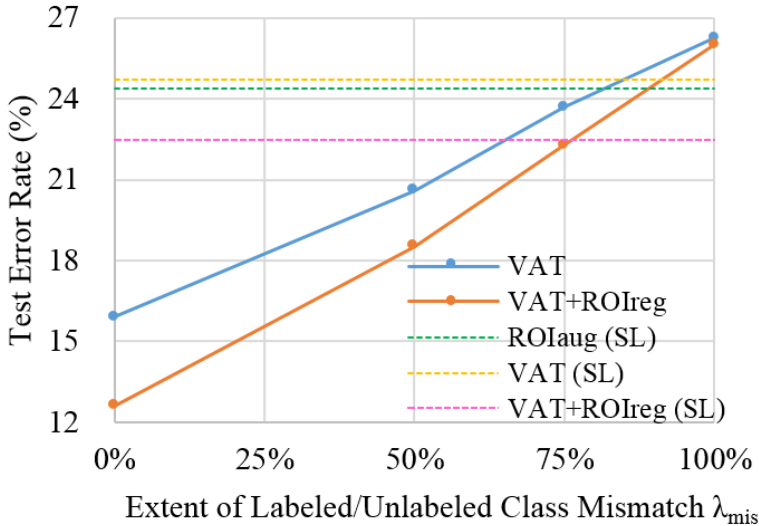}
  \caption{Graph of the results shown in Table \ref{tab6}.}
  \label{fig2}
\end{figure}

\paragraph{Datasets.}
$\mathcal{D}_\mathrm{animal}$ is a data set consisting of 30,000 training data samples belonging to CIFAR-10 animal classes (bird, cat, deer, dog, frog, horse). $\mathcal{D}_\mathrm{artifact}$ is a data set consisting of 20,000 training data samples belonging to the artifact class (airplane, automobile, ship, truck).\\
\\[-2mm]
The same $\mathcal{D}_\mathrm{L}$ is used in all experiments shown in Table \ref{tab6}. Here, $\mathcal{D}_\mathrm{L}$ is composed of data samples randomly sampled 400 per class from $\mathcal{D}_\mathrm{animal}$. Therefore, $\mathcal{D}_\mathrm{L}$ contains 2,400 data samples. The contamination rate $\lambda_{\mathrm{mis}}$ (\%) indicates the extent of labeled/unlabeled class mismatch. That is, $(1 - 0.01 \cdot \lambda_{\mathrm{mis}}) \cdot 20000$ data samples are randomly selected from $\mathcal{D}_\mathrm{animal}$, and $ 0.01 \cdot \lambda_{\mathrm{mis}} \cdot 20000$ data samples are randomly selected from $\mathcal{D}_\mathrm{artifact}$. Then, all these data samples are combined to construct $\mathcal{D}_\mathrm{UL}$. $\mathcal{D}_\mathrm{UL}$ differs for each experiment shown in Table \ref{tab6}.
%
\paragraph{Data Augmentation.}
We use random horizontal flipping and random translation by up to 2 pixels as data augmentation, except in the case of ROIaug. For ROIaug, we use random RGB shuffling, random horizontal flipping,  random translation by up to 4 pixels and Gaussian input noise. Standard supervised learning, ROIaug, VAT, and ROIreg are applied to the data samples obtained by data augmentation.
%
\paragraph{Results.}
The experimental results are shown in Table \ref{tab6} and Figure \ref{fig2}. If only data samples included in $\mathcal{D}_\mathrm{L}$ are used, VAT+ROIreg (SL) achieves the lowest test error rate. In SemiSL, which uses the data samples contained in $\mathcal{D}_\mathrm{L} \cup \mathcal{D}_\mathrm{UL}$, VAT+ROIreg outperforms VAT for all contamination rates. However, the difference decreases with the increase of the contamination rate $\lambda_{\mathrm{mis}}$. From this result, it can be judged that ROIreg is more susceptible to the contamination of $\mathcal{D}_\mathrm{UL}$ with data samples that do not belong to the class to be classified than VAT. Also, at 75\% contamination rate, VAT+ROIreg drops to almost the same performance as VAT+ROIreg(SL). Therefore, VAT+ ROIreg loses its effectiveness as SemiSL at a contamination rate of around 75\%.\\
\\[-2mm]
When the contamination rate is 100\%, it can be said that some representation learning should be applied rather than applying SemiSL. For example, the application of self-supervised learning (SelfSL) (e.g. [4, 19, 13]) may be reasonable. In SelfSL, one input data sample is split into two and a function is defined to evaluate their interrelationship. SelfSL realizes representation learning by minimizing the function. Therefore, the learning method that works as SemiSL in the range where $\lambda_{\mathrm{mis}}$ is small and as SelfSL in the range where $\lambda_{\mathrm{mis}}$ is large can be a candidate for a robust learning method against class distribution mismatch. However, since $\lambda_{\mathrm{mis}}$ is unknown, it is necessary to devise to realize such a learning method.
\newpage
\begin{table}[H]
  \centering
\begin{tabular}{lcccc}
\hline
\multicolumn{1}{c}{\multirow{2}{*}{Method}}     & \multicolumn{4}{c}{$\lambda_\mathrm{mis}$ (\%)}                              \\ \cline{2-5} 
\multicolumn{1}{c}{}                            & 0               & 50              & 75              & 100             \\ \hline
\multicolumn{5}{l}{SemiSL ( $\mathcal{D}_\mathrm{USL} = \mathcal{D}_\mathrm{L} \cup \mathcal{D}_\mathrm{UL}$)}                                 \\ \hline
VAT                                             & 15.90 $\pm$ 0.15    & 20.58 $\pm$ 0.64    & 23.71 $\pm$ 1.02    & 26.29 $\pm$ 0.28    \\
VAT+ROIreg                                      & 12.60 $\pm$ 0.42    & 18.53 $\pm$ 1.13    & 22.28 $\pm$ 1.16    & 26.00 $\pm$ 1.52    \\ \hline
\multicolumn{5}{l}{Supervised learning (SL) \;  ($\mathcal{D}_\mathrm{USL} = \mathcal{D}_\mathrm{L}$ and $\chi^{\mathrm{mb}}_\mathrm{UL} = \chi^{\mathrm{mb}}_\mathrm{L}$ )} \\ \hline
ROIaug(SL)                                      & \multicolumn{4}{c}{24.40 $\pm$ 0.34}                                      \\
VAT(SL)                                         & \multicolumn{4}{c}{24.71 $\pm$ 0.35}                                      \\
VAT+ROIreg (SL)                                 & \multicolumn{4}{c}{22.47 $\pm$ 0.58}                                      \\ \hline
\end{tabular}
\vspace{5mm}
  \caption{Test error rates (\%) on CIFAR-10 (six animal classes) with a varying $\lambda_\mathrm{mis}$. Our experimental results show error4 (see section \ref{sec4.3}). Error bars correspond to the standard deviation over 5 runs. For VAT and VAT+ROIreg, $(m_\mathrm{L},m_\mathrm{UL} )=(64,96)$ and Adam (0.001, 120000, 16000) are applied. For VAT(SL) and VAT+ROIreg(SL), $m_\mathrm{L}=100$ and Adam (0.0001, 24000, 12000) are applied. For ROIaug(SL), $m_\mathrm{L}=100$ and Adam (0.0002, 96000, 48000) are applied. The same $\mathcal{D}_\mathrm{L}$ is used in all experiments.}
  \label{tab6}
\end{table}
\subsection{Effects of weight normalization on CIFAR-10}\label{sec5.3}
In the CIFAR-10 experiments in VAdD \cite{18} and fast-SWA \cite{1}, weight normalization \cite{22} is applied to the convolution layers and the fully connected layers of CNN. Weight normalization has been reported to improve the generalization performance of CNN for CIFAR-10 \cite{22}. For this reason, we also perform experiments in the case of applying weight normalization to the nine convolution layers and one fully connected layer of CNN described in appendix \ref{appA}. Note that mean-only batch normalization \cite{22} is not used, and normal batch normalization is used as in section \ref{sec4}. Also, hyperparameter settings are the same as in Table \ref{tab1}. However, for the learning schedule, Adam (0.00047, 200000, 16000) is applied to CIFAR-10+. For CIFAR-10, Adam (0.001, 200000, 16000) is applied as in section \ref{sec4}. In addition, error2 is adopted as the test error rate in both cases of CIFAR-10+ and CIFAR-10.\\
\\[-2mm]
Table \ref{tab2} shows the experimental results. Both test error rates for CIFAR-10+ and CIFAR-10 improve. As a result, for CIFAR-10+, VAT+ROIreg+ENT outperforms the test error rate of 9.22\% achieved by VAdD (QE)+VAT+ENT. Therefore, it can be said that Weight Normalizaton is effective for ROIreg.\\
\\[-2mm]
However, VAT+ROIreg+ENT does not reach the result of 9.05\% achieved by MT+Fast-SWA. On the other hand, VAdD is a learning method that can complement each other with VAT. In fact, for CIFAR-10+, VAdD(QE)+VAT+ENT improves the result of 11.96\% achieved by VAT+ENT (implemented by Park et al.) to the result of 9.22\%. Thus, using VAdD(QE) and VAT+ROIreg+ENT simultaneously may improve the current best results for CIFAR-10+.\\
%
\section{Conclusions}\label{sec6}
We proposed ROI regularization (ROIreg) as a new method of semi-supervised learning. VAT+ROIreg+ENT achieved the state-of-the-art performances on SVHN, SVHN+ and CIFAR-10. On the other hand, on CIFAR-10+, this combination achieved a result of 9.13\%. This does not exceed the known best result \cite{1} of 9.05\%. However, using VAT+ROIreg+ENT simultaneously with VAdD (QE) may improve this best result.\\
\\[-2mm]
VAT+ROIreg was also effective for CIFAR-10 (4000 labels) using only labeled data. We also proposed ROI augmentation (ROIaug) as a new method of data augmentation using only labeled data. ROIaug is one of the variations of ROIreg. ROIaug was able to improve the results of supervised learning on both SVHN (1000 labels) and CIFAR-10 (4000 labels).\\
\\[-2mm]
From the point of view of Class Distribution Mismatch, the conventional performance evaluation for SemiSL is performed on a single-point spectrum with zero contamination rate. However, it is also important to evaluate the spectrum with a spread over all contamination rates. We will study the method of SemiSL that exhibits excellent performance in this sense.\\

\bibliographystyle{unsrt}  


\vspace{10mm}
\appendix
\section{CNN used in our experiments}\label{appA}
\begin{table}[H]
\centering
\begin{tabular}{ll} 
\hline
Layer & Hyperparameters \\ 
\hline
Convolution + BN + Leaky ReLU (0.1) & 128 filters, $3 \times 3$ \\
Convolution + BN + Leaky ReLU (0.1) & 128 filters, $3 \times 3$ \\
Convolution + BN + Leaky ReLU (0.1) & 128 filters, $3 \times 3$ \\
Pooling + Dropout ($p = 0.5$) & Maxpool $2 \times 2$, stride 2 \\
Convolution + BN + Leaky ReLU (0.1) & 256 filters, $3 \times 3$ \\
Convolution + BN + Leaky ReLU (0.1) & 256 filters, $3 \times 3$ \\
Convolution + BN + Leaky ReLU (0.1) & 256 filters, $3 \times 3$ \\
Pooling + Dropout ($p = 0.5$) & Maxpool $2 \times 2$, stride 2 \\
Convolution + BN + Leaky ReLU (0.1) & 512 filters, $3 \times 3$ \\
Convolution + BN + Leaky ReLU (0.1) & 256 filters, $1 \times 1$ \\
Convolution + BN + Leaky ReLU (0.1) & 128 filters, $1 \times 1$ \\
Pooling & Global average pooling ($6 \times 6 \rightarrow 1 \times 1$) \\
Fully connected + BN${}^a$ + Softmax & $128  \rightarrow  10$ \\
\hline               
\end{tabular}
\vspace{5mm}
  \caption{The convolutional network architecture used in our experiments. BN refers to batch normalization using the mean and standard deviation on each minibatch. ${}^a$ Not applied on CIFAR-10 experiments.}
  \label{tab7}
\end{table}

\end{document}